\pdfoutput=1
\documentclass{article}

\usepackage[preprint]{neurips_2026}
\usepackage[utf8]{inputenc} % allow utf-8 input
\usepackage[T1]{fontenc}    % use 8-bit T1 fonts
\usepackage{hyperref}       % hyperlinks
\usepackage{url}            % simple URL typesetting
\usepackage{booktabs}       % professional-quality tables
\usepackage{amsfonts}       % blackboard math symbols
\usepackage{amsmath}        % math environments
\usepackage{amssymb}        % math symbols
\usepackage{nicefrac}       % compact symbols for 1/2, etc.
\usepackage{microtype}      % microtypography
\usepackage{xcolor}         % colors
\usepackage{graphicx}       % figures
\usepackage{verbatim}       % \verbatiminput
\usepackage{placeins}       % \FloatBarrier
\usepackage{subcaption}     % side-by-side subfigures
\usepackage{float}          % [H] placement specifier
\usepackage{multirow}       % multi-row cells
\usepackage{alltt}          % verbatim-like env with inline LaTeX support (for bolding rows)

\title{Weak-to-Strong Elicitation \\ via Mismatched Wrong Drafts}

\author{%
  Wei Deng\thanks{Correspondence: \texttt{wdeng@wustl.edu}.} \\
  Independent Researcher \\
  {\footnotesize Data, code, and models: \url{https://github.com/weiddeng/mismatched-wrong-drafts}}
}

\begin{document}

\maketitle

\vspace{-8pt}% reclaim the vertical space the author-block URL line adds, restoring the v1 page-1 abstract split (Contributions atop p.3)
\begin{figure}[H]
\centering
\includegraphics[width=\textwidth]{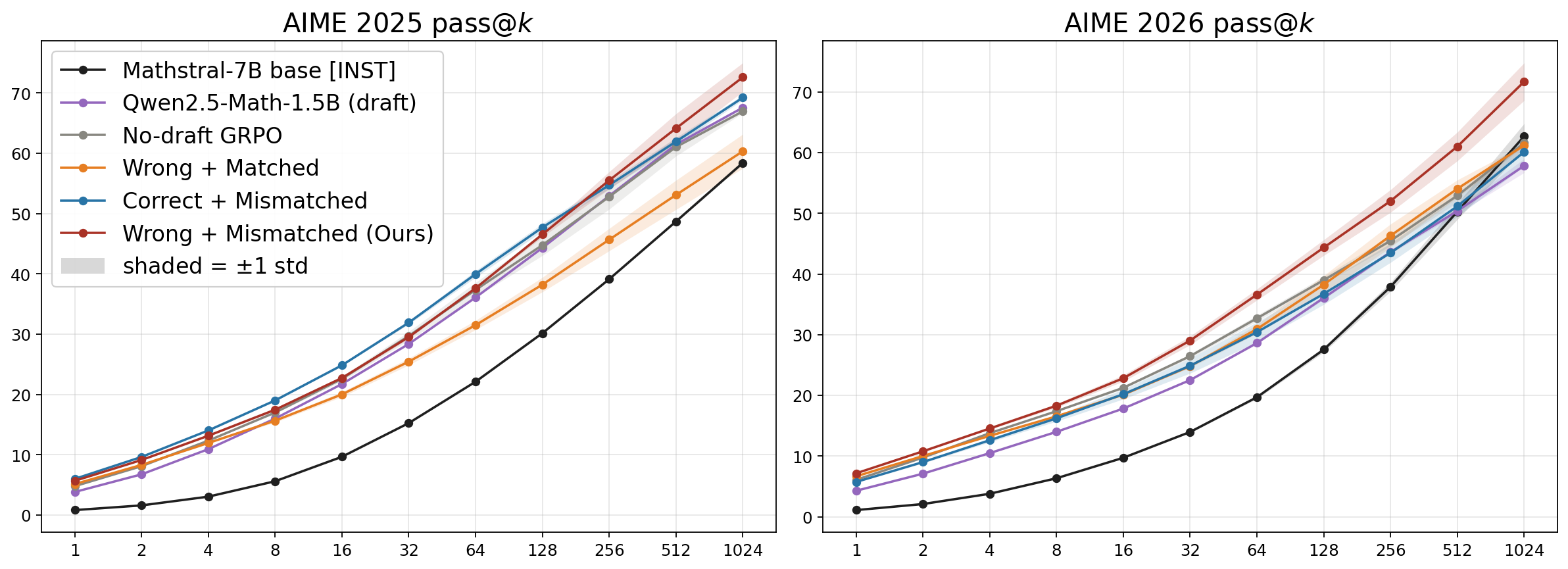}
\caption{On out-of-distribution AIME 2025 and 2026, the mismatched-wrong variant (ours, red) uniquely lifts pass@$k$ above both Mathstral-7B and the Qwen2.5-Math-1.5B draft model at every sample budget from $k{=}1$ to $k{=}1024$. Mathstral-7B is evaluated in its native \texttt{[INST]} chat format; all other variants use the training-matched nodraft prompt (literal \texttt{N/A} placeholder). $N{=}2048$ samples per problem at $T{=}0.6$, top-$p{=}0.95$, max 4096 completion tokens; mean across 2 seeds $s{=}\{42,137\}$.}
\label{fig:aime-passk}
\end{figure}

\begin{abstract}
We consider whether off-policy experience from a smaller, weaker model can elicit capability in a stronger learner that on-policy RL fine-tuning (e.g., GRPO) does not reach. We find that injecting mathematically \emph{wrong} drafts from a smaller but more domain-trained model---\emph{mismatched} to the current problem---into a stronger learner's GRPO context consistently outperforms standard on-policy GRPO on held-out MATH-500 and out-of-distribution AIME 2025/2026. Concretely, we use Mathstral-7B as the learner, Qwen2.5-Math-1.5B as the draft model, 8.8K Level 3--5 MATH problems (with MATH-500 held out), and train with Dr.~GRPO. Mismatch is an active ingredient: shuffling drafts to mismatched problems while holding everything else constant yields $+1.62$pp on MATH-500 (greedy pass@1) over the matched-wrong variant ($n{=}10$ seeds, $p{=}0.0015$, Welch's $t$). In fact, the mismatched-wrong variant leads all other variants we tested on MATH-500 across both greedy pass@1 and sampling pass@$k$. On out-of-distribution AIME 2025 and 2026, the mismatched-wrong variant uniquely lifts pass@$k$ above both Mathstral-7B (in its native [INST] format) and the Qwen2.5-Math-1.5B draft model at every sample budget from $k{=}1$ to $k{=}1024$ across 2 seeds ($+14.2$pp on 2025 and $+9.0$pp on 2026 at pass@1024 over Mathstral-7B), and at pass@1024 also leads no-draft, matched-wrong, and mismatched-correct variants on both years. All variants use the same prompt with no draft injection at test time. The recipe---trained on a single GPU with no SFT, no reward models, no synthesized data, and no produce-critique-revise inner loop---reaches $71.98\%$ MATH-500 on Mathstral-7B-v0.1, the highest published result on this model to our knowledge, surpassing the heavier WizardMath pipeline at $70.9\%$ on full MATH (SFT + PPO with process/instruction reward models).
\end{abstract}

\section{Introduction}
\label{sec:introduction}

Several paradigms aim to improve large language model reasoning: supervised fine-tuning on \emph{correct} traces, either from stronger models (e.g., DeepSeek-R1-Distill-Qwen~\citep{deepseek2025r1}) or self-bootstrapped from the model's own correct rollouts (STaR~\citep{zelikman2022star}, \citet{huang2023llmselfimprove}); iterative correction-and-refinement pipelines that produce, critique, and revise their own outputs~\citep{madaan2023selfrefine}, including RL-trained self-correction~\citep{kumar2025score}; reinforcement learning from human feedback (RLHF~\citep{ouyang2022instructgpt}), which trains against a learned reward model fitted on human preferences; and on-policy reinforcement learning with verifiable rewards (RLVR), most prominently GRPO~\citep{shao2024deepseekmath}, which trains on the model's own rollouts using a verifier. On-policy RL is appealing because it does not require any supervision except a verifier, but in its standard form the input distribution is narrow: each training prompt is the bare problem statement, and reward can only select among trajectories the strong model already samples in response.
This is a recognized limitation: a growing line of empirical analyses argues that on-policy RL fine-tuning sharpens existing modes rather than expanding the base model's intrinsic coverage, with pass@$k$ at large $k$ often matching or falling below the base~\citep{yue2025passk}.

A natural way to expand what the learner produces under GRPO rollouts---and therefore what reward can score and select---is to broaden the training prompt distribution, while keeping the learner robust under the resulting training--inference distribution discrepancy. Consider another model that has been more domain-trained: it has seen more data, accumulated a record of attempts, mistakes, and partial solutions that are uncharted and lie dormant in the learner. We focus on the special case where the other model is \emph{smaller}, with different training experience from the learner, and ask whether its \emph{wrong} draft traces, placed in the learner's prompt context window, can elicit capability that on-policy GRPO from bare prompts does not reach.

The answer hinges on a second choice: whether the injected draft is about the current problem or about a different one. With everything else fixed---learner (Mathstral-7B), draft model (Qwen2.5-Math-1.5B), data ($\sim$8.8K Level 3--5 MATH problems with MATH-500 held out), algorithm (Dr.~GRPO~\citep{liu2025drgrpo}), eval protocol---we isolate two axes simultaneously: draft content (correct vs.\ wrong) and draft assignment (matched vs.\ mismatched).
We compare these four variants and a no-draft GRPO baseline, as well as the Mathstral-7B base, on MATH-500 and out-of-distribution AIME 2025/2026. Only mismatched-wrong consistently exceeds no-draft GRPO on both evaluations and uniquely lifts pass@$k$ above both Mathstral-7B and the Qwen2.5-Math-1.5B draft model at every sample budget from $k{=}1$ to $k{=}1024$ on AIME 2025/2026 (Figure~\ref{fig:aime-passk}).

Both the mismatch step and the wrongness of the draft are active ingredients. We randomly select a draft with a wrong answer (avoiding wrong-but-quasi-correct drafts when possible), and shuffle it to a different problem; the draft, now about a different problem, implicitly \emph{lifts} the training prompt to a more general but \emph{masked} task, of which the original bare problem is a degenerate special case. The mismatched wrong draft is an \emph{observation}---an off-policy trace of an attempt at a masked problem, sitting in context alongside the actual question. The strong model produces a solution from scratch in a single rollout per prompt, with no produce-critique-revise loop or second pass. The recipe is standard on-policy RL fine-tuning. GRPO's reward then selects, across rollouts, the solutions the strong model finds from its own intrinsic capabilities. Because the task of interest is a degenerate special case of the training prompt, the training--inference discrepancy is minimal. The weak model is not supervised-fine-tuning the strong learner~\citep{burns2023w2s}, and the strong learner is not correcting the weaker draft.

The recipe is materially simpler than the strongest published Mathstral-7B-v0.1 pipeline yet beats it: with a single GPU, no SFT, no reward models, no synthesized data, and no produce-critique-revise inner loop, the mismatched-wrong variant reaches \(71.98\%\) on MATH-500 (\(n{=}10\) seeds, 95\% CI \(\pm 0.80\)pp). For reference, WizardMath~\citep{luo2025wizardmath} reports 70.9\% on full MATH using a synthesized SFT stage followed by PPO with both a process and an instruction reward model.

\paragraph{Contributions.}
\begin{itemize}
    \item \textbf{Weak-to-strong elicitation can simultaneously sharpen and expand the strong learner's coverage under on-policy RLVR with GRPO.} Recent analyses argue that on-policy RL fine-tuning only sharpens existing modes. Our recipe is a counterexample: MATH-500 greedy pass@1 lifts by $+17.78$pp over Mathstral-7B base ($n{=}10$ seeds, $p<0.0001$) and pass@$k$ on out-of-distribution AIME 2025/2026 lifts above Mathstral-7B base at every sample budget from $k{=}1$ to $k{=}1024$ (2 seeds).
    \item \textbf{We show that mismatch \(\times\) wrongness is the active ingredient.} We isolate the full \(2{\times}2\) (draft assignment matched/mismatched \(\times\) draft content correct/wrong) variants under the same draft model, training data, and recipe; only the mismatched-wrong variant consistently lifts above the Mathstral-7B base.
    \item \textbf{A small recipe that beats heavier pipelines on Mathstral-7B.} \(71.98\%\) on MATH-500---exceeding WizardMath's heavier 70.9\% (full MATH)---with a single-GPU and outcome-reward-only recipe.
\end{itemize}

\section{Related Work}
\label{sec:related-work}

\textbf{RLVR for mathematical reasoning.} Reinforcement learning has driven much of the recent progress in LLM for mathematics, exemplified by GRPO and descendants~\citep{shao2024deepseekmath,liu2025drgrpo,yu2025dapo} and the ``zero''-style line of work showing that strong reasoning emerges directly from RL without an SFT stage~\citep{deepseek2025r1,hu2025openreasonerzero,zeng2025simplerlzoo}. WizardMath~\citep{luo2025wizardmath} represents the heavier end of the spectrum, combining synthesized SFT data with PPO and process/instruction reward models; it is our headline $70.9\%$ Mathstral-7B comparison. Our recipe uses Dr.~GRPO~\citep{liu2025drgrpo} unchanged, and the novelty sits at the \emph{task} the learner is trained on.

\textbf{Coverage vs.\ sharpening under RL post-training.} A growing line of empirical analyses argues that on-policy RL fine-tuning sharpens existing modes while leaving the base model's pass@$k$ coverage at large $k$ unchanged or even reduced~\citep{yue2025passk}; concurrently, methods that explicitly trade off generation diversity against quality during RL have been proposed~\citep{li2025diversequality}. Our recipe is a counterexample to the sharpen-only reading (see \S\ref{sec:experiments}).

\textbf{Weak-to-strong and self-improvement.} Prior approaches all use the weaker (or earlier) model as a supervision signal: weak-to-strong supervision distills a weaker model's labels into a stronger one~\citep{burns2023w2s}; self-bootstrapping methods iteratively retrain on the model's own correct rollouts filtered by reward (STaR~\citep{zelikman2022star}, ReST$^{EM}$~\citep{singh2024restem}); iterative correction-and-refinement pipelines train models to revise their own attempts via produce--critique--revise loops~\citep{welleck2023selfcorrect}, and SCoRe~\citep{kumar2025score} uses multi-turn RL and reward shaping to train models to correct their own first-attempt mistakes. Closest to our setting, both \citet{burns2023w2s} and \citet{bansal2025smallerweakerbetter} use a weaker model to produce supervised training data for a stronger one (labels and synthesized data respectively); we instead inject wrong drafts into the strong model's GRPO context window. In all of these prior approaches the weaker (or earlier) model serves as a teacher or starting point for revision; in ours, it is an off-policy explorer that lifts the training task to a more general one, while the loss remains on-policy with respect to the strong learner.

\section{Method}
\label{sec:method}

\subsection{Data}
\label{sec:data}

Training uses $\sim$8.8K Level 3--5 problems among the 12K problems in MATH~\citep{hendrycks2021math} after removing the 500 problems of MATH-500~\citep{lightman2024letsverify}. All training runs shuffle this set with the same random seed and see identical problems in identical order at every step. Testing uses the held-out MATH-500 and AIME 2024/2025/2026~\citep{matharena}.

\subsection{Wrong Drafts}
\label{sec:wrong-drafts}

For each training problem $x$, we sample 32 draft completions from the weaker model $\pi_W$ at temperature $T{=}0.8$, top-$p{=}0.95$, max 2560 completion tokens. We define a helper \texttt{mathematically\_quasi\_correct}$(\cdot)$ that runs math-verify~\citep{kydlicek2025mathverify} against an answer extracted via a prioritized fallback chain: \texttt{\textbackslash boxed\{$\cdot$\}} first, then natural-language patterns (``the answer is X''), inline math expressions ($\$\ldots\$$), and bare assignment lines (``var = VALUE''). Among the 32, we randomly sample a completion that is wrong and non-trivially so (\texttt{mathematically\_quasi\_correct}=\texttt{False}), falling back to one rejected by the strict boxed-only criterion if all are quasi-correct, and finally to any completion. The result is an offline paired set $\{(x, d^-_x)\}_{x \in \mathcal{D}}$ with $\sim$8.8K problems, each carrying one selected draft, sampled once before RL training begins.

\subsection{Mismatched Wrong Drafts}
\label{sec:approach}

We apply a random 1-1 derangement $\sigma : \mathcal{D} \to \mathcal{D}$, pairing each problem with the wrong draft of another problem:
\begin{equation}
\mathrm{train\;dataset} \;=\; \{(x, \, d^-_{\sigma(x)}) : x \in \mathcal{D}\}.
\end{equation}
Concretely, we draw a random permutation. Note that in an unconstrained random permutation the expected number of fixed points is 1. We then minimally swap away any fixed points, still keeping a valid permutation but now with $\sigma(x) \neq x$ for all $x$. We run on-policy Dr.~GRPO~\citep{liu2025drgrpo} on $\pi_S$ over augmented prompts $\tilde{x} = \mathrm{Template}(x, d^-_{\sigma(x)})$; rollouts and gradients remain on-policy with respect to $\pi_S$. The exact prompt template is shown in Figure~\ref{fig:prompt-template}. The derangement is fixed once at the start of training.

\begin{figure}[t]
\centering
\fbox{\begin{minipage}{0.92\textwidth}
\small\ttfamily
Problem: \{problem\}\\[4pt]
Thinking: \{draft\}\\[4pt]
The thinking section may contain errors. Solve the math problem step by step. Write your own correct solution. Put your final answer within \textbackslash boxed\{\}.\\[4pt]
Correct Solution:
\end{minipage}}
\caption{Prompt template. At training time, \texttt{\{draft\}} is the (mismatched, wrong) draft $d^-_{\sigma(x)}$. At evaluation time, \texttt{\{draft\}} is the literal string ``\texttt{N/A}''.}
\label{fig:prompt-template}
\end{figure}

\subsection{Reward}
\label{sec:reward}

The reward is binary and outcome-only: $1$ if \texttt{mathematically\_quasi\_correct(completion, gold)} returns True, and $0$ otherwise. We opt for this lenient check rather than a strict boxed-only requirement to accelerate reward signal acquisition during training. We use no format, length, or process reward. We apply Dr.~GRPO~\citep{liu2025drgrpo} to maximize the efficiency of our limited completion-length budget. Training details are in \S\ref{sec:exp-setup}.

\section{Experiments}
\label{sec:experiments}

\subsection{Setup}
\label{sec:exp-setup}

\textbf{Training.} We fine-tune Mathstral-7B~\citep{mistral2024mathstral} via LoRA adapters of rank $16$ on all 7 linear projections per transformer block (attention + MLP)~\citep{hu2022lora}, drawing drafts from Qwen2.5-Math-1.5B~\citep{yang2024qwen25math}, on a single B200 GPU. Optimizer: AdamW with constant learning rate $5\times 10^{-6}$, $\beta_2{=}0.99$. RL config (Dr.~GRPO): $\beta{=}0$ (no KL penalty), group size $G{=}16$, gradient accumulation $4$, 2222 steps (1 epoch). Generation: max completion length $4096$ tokens, max prompt length $3072$. Checkpoints saved every $50$ steps. Each run takes up to 30+ hours wall-clock. Implementation uses TRL~\citep{vonwerra2022trl}, vLLM~\citep{kwon2023vllm}, and Unsloth~\citep{unsloth}.

\textbf{Evaluation.} Our evaluation spans the MATH-500 and AIME 2024--2026 datasets, tracking two primary metrics: (1) greedy pass@1 ($T{=}0$, max 4096 completion tokens) and (2) sampling pass@$k$ across various budgets ($N=256$ samples per problem for MATH-500, $N=2048$ for AIME, max 4096 completion tokens, $T=0.6$, top-$p=0.95$), calculated via the unbiased estimator from~\citet{chen2021codex}. We maintain a consistent prompt template (Figure~\ref{fig:prompt-template}) during evaluation for our trained models and the Qwen2.5-Math-1.5B drafter. Specifically, the \texttt{\{draft\}} field is populated with the literal string ``\texttt{N/A}'' during evaluations (as well as during training of the no-draft variant). The only exception is the Mathstral-7B base model, which we test using its default \texttt{[INST]} chat format: \texttt{\{problem\}\textbackslash n\textbackslash n} followed by ``Please reason step by step, and put your final answer within \texttt{\textbackslash boxed\{\}}.'', all enclosed in \texttt{[INST]}\,\ldots\,\texttt{[/INST]} tokens. Section~\ref{sec:exp-aime-clean} confirms that our performance gains over Mathstral-7B remain valid despite this formatting difference.

\subsection{MATH-500}
\label{sec:exp-main}

\begin{figure}[t]
\centering
\includegraphics[width=\textwidth]{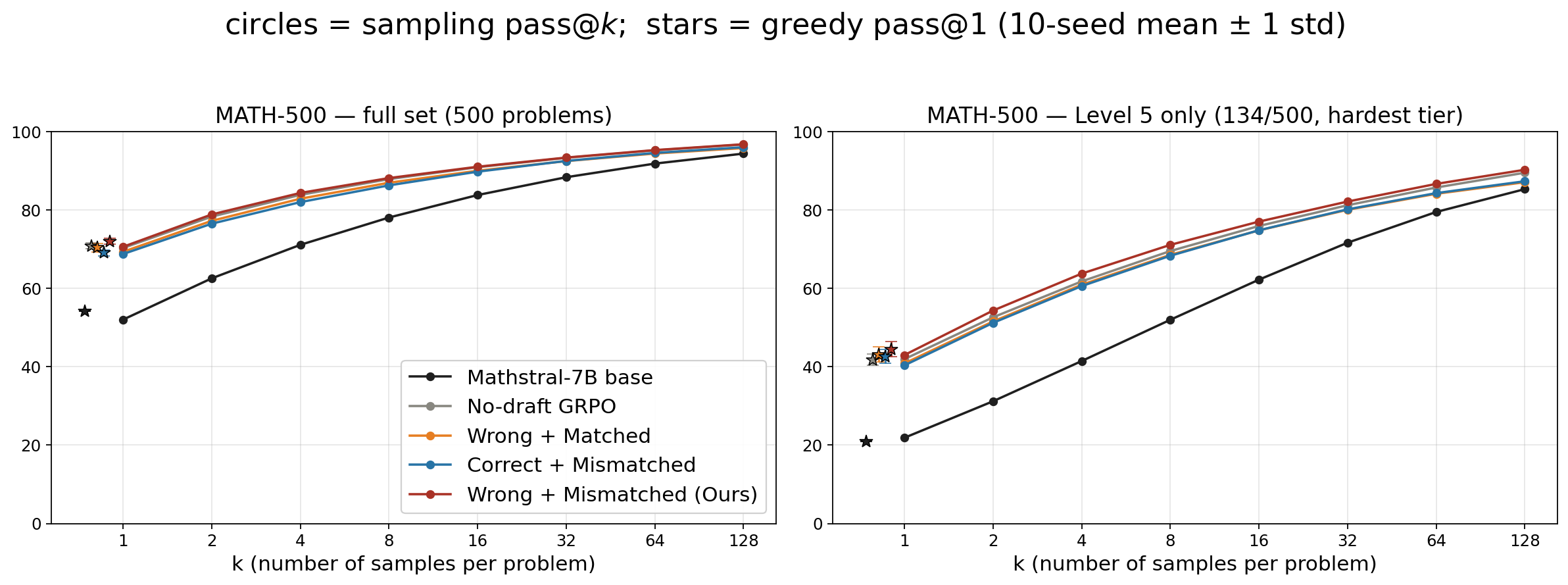}
\caption{MATH-500 pass@$k$. \textbf{Left:} overall (500 problems). \textbf{Right:} Level 5 only (134 problems). 2-seed mean ($s{=}\{42, 137\}$) for all 5 lines (Mathstral-7B base, no-draft GRPO, matched-wrong, mismatched-correct, mismatched-wrong).}
\label{fig:math500-passk-overall-L5}
\end{figure}

The mismatched-wrong variant achieves $71.98\%$ on MATH-500 ($n=10$ seeds, 95\% CI $\pm 0.80$pp), surpassing the heavier WizardMath pipeline at $70.9\%$ on full MATH (Table~\ref{tab:setup-a}). Beyond greedy pass@1, the mismatched-wrong variant also leads on sampling pass@$k$ (Figure~\ref{fig:math500-passk-overall-L5}).

\begin{table}[!htbp]
  \caption{Greedy pass@1 on MATH for methods fine-tuning Mathstral-7B-v0.1. WizardMath reports on the full MATH test (5{,}000 problems); our results on the MATH-500 subset~\citep{lightman2024letsverify}.}
  \label{tab:setup-a}
  \centering
  \small
  \begin{tabular}{lcc}
    \toprule
    Work & MATH (full) & MATH-500 \\
    \midrule
    WizardMath~\citep{luo2025wizardmath} & 70.9 & --- \\
    \midrule
    \textbf{Mismatched-wrong (ours)} & --- & $\mathbf{71.98}$ \\
    \bottomrule
  \end{tabular}
\end{table}

\FloatBarrier

\subsection{AIME 2025 and 2026}
\label{sec:exp-aime-clean}

\begin{table}[!htbp]
  \caption{Mean across 2 seeds ($s{=}\{42, 137\}$). Endpoint values of Figure~\ref{fig:aime-passk}.}
  \label{tab:aime-passk}
  \centering
  \small
  \begin{tabular}{lrrrr}
    \toprule
    & \multicolumn{2}{c}{AIME 2025 (n=30)} & \multicolumn{2}{c}{AIME 2026 (n=30)} \\
    \cmidrule(r){2-3}\cmidrule(r){4-5}
     & pass@1 & pass@1024 & pass@1 & pass@1024 \\
    \midrule
    Mathstral-7B base ([INST] prompt)    & 0.81 & 58.40 & 1.12 & 62.69 \\
    Mathstral-7B base (nodraft prompt)   & 0.87 & 64.21 & 1.16 & 55.47 \\
    Qwen2.5-Math-1.5B (draft model) & 3.82 & 67.57 & 4.30 & 57.84 \\
    No-draft GRPO ([INST])        & 3.72 & 65.49 & 4.29 & 63.33 \\
    No-draft GRPO (nodraft)       & 4.78 & 66.94 & 6.09 & 61.58 \\
    Matched-wrong ([INST])        & 2.53 & 64.62 & 3.01 & 55.15 \\
    Matched-wrong (nodraft)       & 5.13 & 60.33 & \underline{6.68} & 61.20 \\
    Mismatched-correct ([INST])   & 3.57 & 68.19 & 4.12 & 59.12 \\
    Mismatched-correct (nodraft)  & \textbf{5.99} & \underline{69.22} & 5.75 & 60.11 \\
    Mismatched-wrong (ours, [INST])  & 3.37 & 60.92 & 4.50 & \underline{65.73} \\
    \textbf{Mismatched-wrong (ours, nodraft)} & \underline{5.66} & $\mathbf{72.60}$ & $\mathbf{7.17}$ & $\mathbf{71.68}$ \\
    \bottomrule
  \end{tabular}
\end{table}

If the recipe merely sharpens the strong model's distribution---reweighting probability mass toward already-reachable solutions---its pass@$k$ curve at large $k$ should saturate at or below the base model. If it expands the policy's reachable set, the curve should dominate the baseline at every $k$. We probe this on out-of-distribution AIME 2025 and AIME 2026, where contamination of the underlying models is implausible (both years post-date the training cutoff of Mathstral-7B and Qwen2.5-Math-1.5B). The data falls on the side of expansion (Figure~\ref{fig:aime-passk}, Table~\ref{tab:aime-passk}): $+14.2$pp on 2025 and $+9.0$pp on 2026 at $k{=}1024$ over Mathstral-7B in its native \texttt{[INST]} format. Table~\ref{tab:aime-passk} reports both prompting formats for completeness; within each model the two formats give comparable numbers, but the training-consistent format (\texttt{[INST]} for base, \texttt{nodraft} for the trained variants) generally does better.

\begin{figure}[!htbp]
\centering
\includegraphics[width=\textwidth]{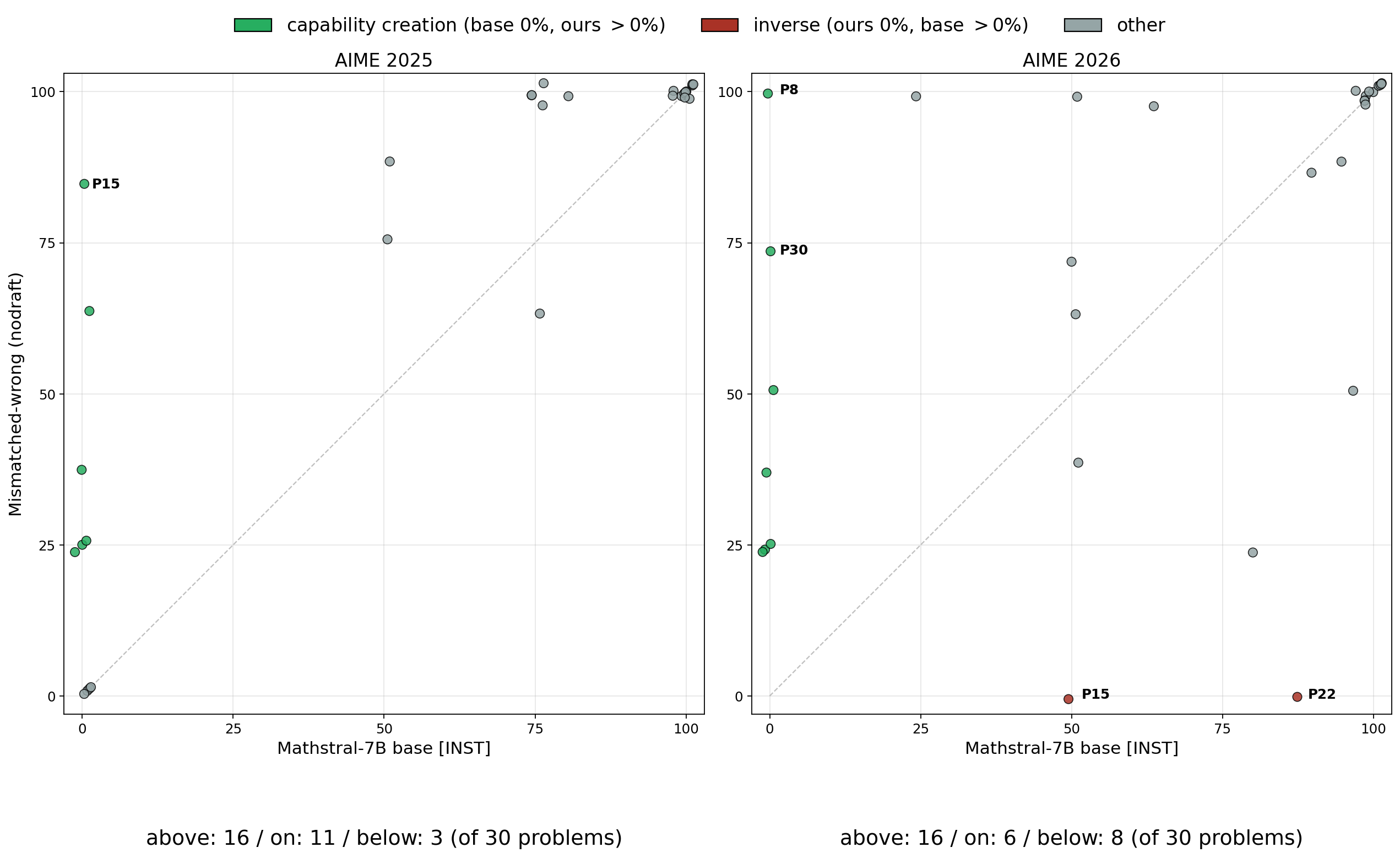}
\caption{Per-problem AIME pass@1024 (one dot = one problem, 2-seed mean, slight diagonal jitter): mismatched-wrong (ours, nodraft) vs Mathstral-7B base (\texttt{[INST]}). Points above the diagonal indicate our variant wins. Green: capability-creation cases (base $0\%$, ours $>0\%$; $13$ problems). Red: inverse (ours $0\%$, base $>0\%$; $2$ problems).}
\label{fig:aime-per-problem}
\end{figure}

\textbf{Per-problem analysis.} The overall improvement stems from large, concentrated gains on specific problems rather than marginal improvements across the board. Furthermore, these gains outweigh the losses in both frequency and magnitude. Out of 60 AIME 2025+2026 problems, $11$ see a pass@1024 increase of $\geq 30$pp over the Mathstral-7B baseline, compared to only $4$ that lose $\geq 30$pp. Similarly, $6$ problems gain $\geq 50$pp while only $3$ lose $\geq 50$pp (Figure~\ref{fig:aime-per-problem}).

We also observe $13$ ``capability-creation'' cases---instances where the baseline scores $0\%$ but our model achieves a positive pass rate. The most striking of these reach near-perfect success (e.g., AIME 2026 P8: $0\%\!\to\!100\%$; AIME 2025 P15: $0\%\!\to\!84.4\%$). Conversely, the inverse scenario---where our model collapses to $0\%$ on a problem the baseline could solve---is rare, occurring on just $2$ problems (AIME 2026 P22: $87.5\%\!\to\!0\%$; AIME 2026 P15: $50\%\!\to\!0\%$).

\FloatBarrier

\subsection{AIME 2024}
\label{sec:exp-aime2024}

AIME 2024 predates both Mathstral-7B-v0.1 and Qwen2.5-Math-1.5B, so one or both may have seen it during training. We thus exclude it from the headline claim and report it here for completeness, with caveats discussed in \S\ref{sec:discussion} and reasoning-rigor results in \S\ref{app:rigor-scan}.

\begin{figure}[!htbp]
\centering
\includegraphics[width=0.78\textwidth]{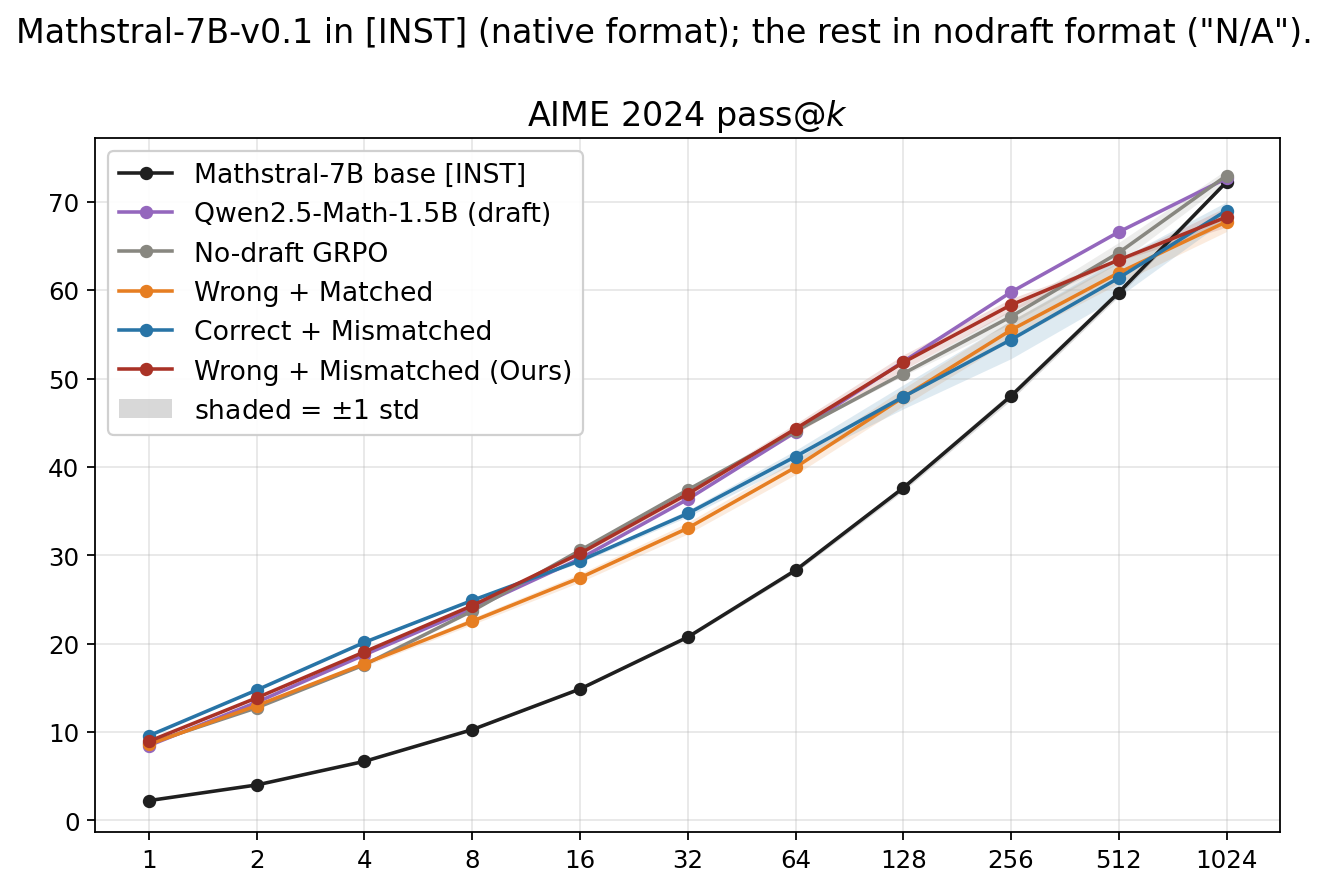}
\caption{AIME 2024 pass@$k$ ($n{=}2048$ samples per problem, mean across 2 seeds $s \in \{42, 137\}$). Base lags trained variants at low $k$ but catches up at $k{=}1024$ to within $\sim\!1$pp of the leaders.}
\label{fig:aime2024-passk}
\end{figure}

\begin{table}[!htbp]
  \caption{Mean across 2 seeds ($s{=}\{42, 137\}$). Endpoint values of Figure~\ref{fig:aime2024-passk}.}
  \label{tab:aime2024-passk}
  \centering
  \small
  \begin{tabular}{lcc}
    \toprule
    Configuration & pass@1 (\%) & pass@1024 (\%) \\
    \midrule
    Mathstral-7B base ([INST])           & 2.22 & 72.25 \\
    Mathstral-7B base (nodraft)          & 2.29 & 66.14 \\
    Qwen2.5-Math-1.5B (draft model)      & 8.45 & \underline{72.65} \\
    No-draft GRPO ([INST])               & 6.97 & 63.68 \\
    No-draft GRPO (nodraft)              & 8.69 & $\mathbf{72.97}$ \\
    Matched-wrong ([INST])               & 4.34 & 68.61 \\
    Matched-wrong (nodraft)              & 8.59 & 67.75 \\
    Mismatched-correct ([INST])          & 6.04 & 64.87 \\
    Mismatched-correct (nodraft)         & $\mathbf{9.57}$ & 69.02 \\
    Mismatched-wrong (ours, [INST])      & 5.62 & 70.36 \\
    \textbf{Mismatched-wrong (ours, nodraft)} & \underline{8.94} & 68.28 \\
    \bottomrule
  \end{tabular}
\end{table}

\begin{figure}[!htbp]
\centering
\begin{minipage}[c]{0.52\textwidth}
\centering
\includegraphics[width=\linewidth]{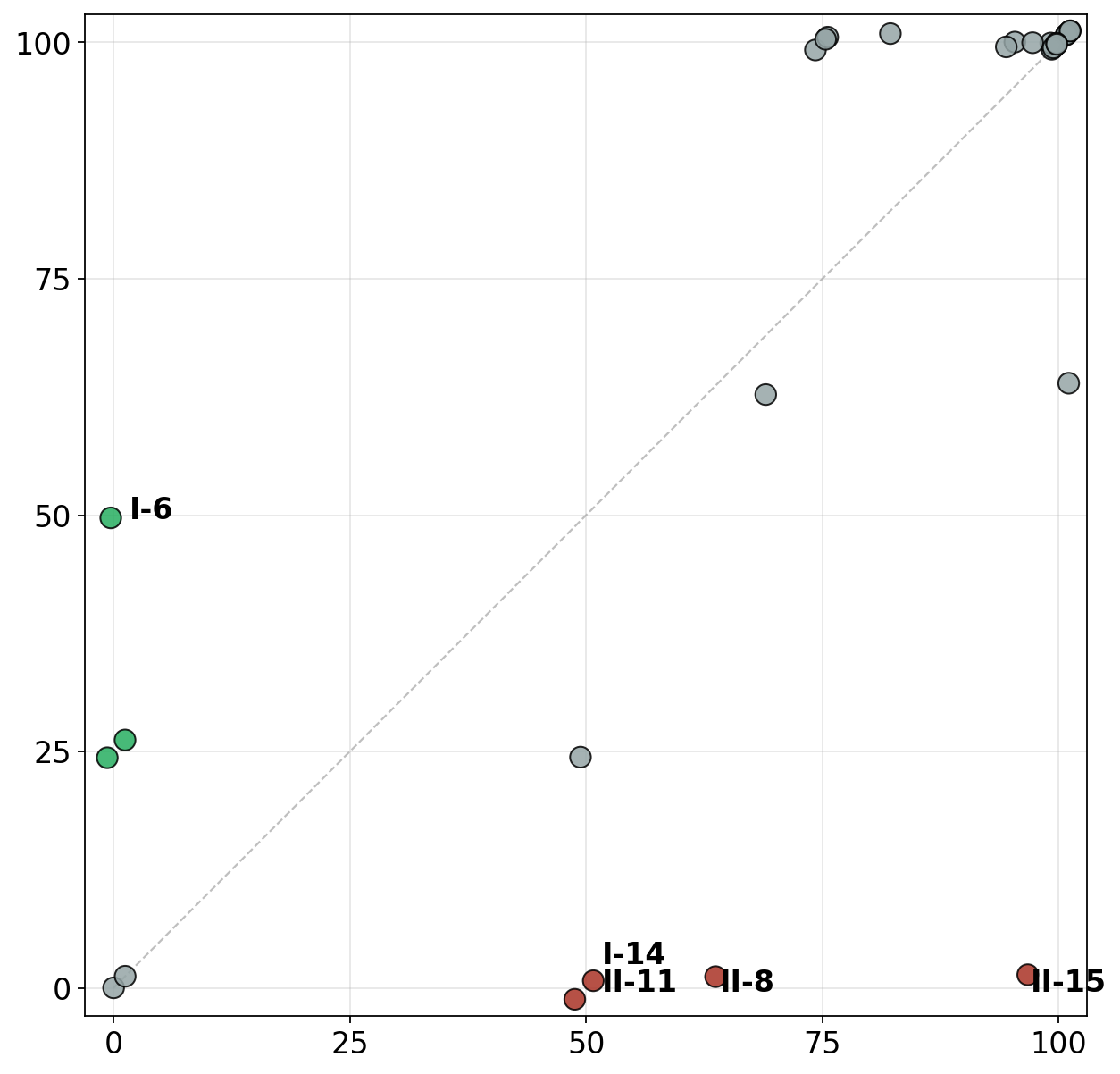}
\end{minipage}\hfill
\begin{minipage}[c]{0.45\textwidth}
\caption{Per-problem AIME 2024 pass@1024 (one dot = one problem, 2-seed mean, slight diagonal jitter): $y$-axis = mismatched-wrong (ours, nodraft), $x$-axis = Mathstral-7B base (\texttt{[INST]}). Of 30 problems: 12 above the diagonal, 11 on (gap $<0.1$pp; most are problems both models solve at $\sim$100\%), 7 below. Green: capability-creation cases (base $0\%$, ours $>0\%$; $3$ problems). Red: inverse (ours $0\%$, base $>0\%$; $4$ problems). The asymmetry observed on AIME 2025/2026 (Figure~\ref{fig:aime-per-problem}) reverses on AIME 2024. Per \S\ref{app:rigor-scan}, however, the creation case I-6 yields a rigorous solution from our variant, while none of the $4$ inverse cases yield any valid solution.}
\label{fig:aime2024-per-problem}
\end{minipage}
\end{figure}

Unlike on 2025/2026, the mismatched-wrong recipe (ours, nodraft) does not lead the other trained variants here: no-draft GRPO is $\sim\!4.7$pp ahead at pass@$1024$ ($72.97\%$ vs $68.28\%$). Within the wrong-draft axis, though, it still leads matched-wrong by $1$--$4$pp across $k$ (Figure~\ref{fig:aime2024-passk}).

\FloatBarrier

\subsection{The 2$\times$2 variants vs.\ no-draft GRPO vs.\ Mathstral-7B}
\label{sec:exp-mismatch-iso}

\begin{figure}[!htbp]
\centering
\includegraphics[width=\textwidth]{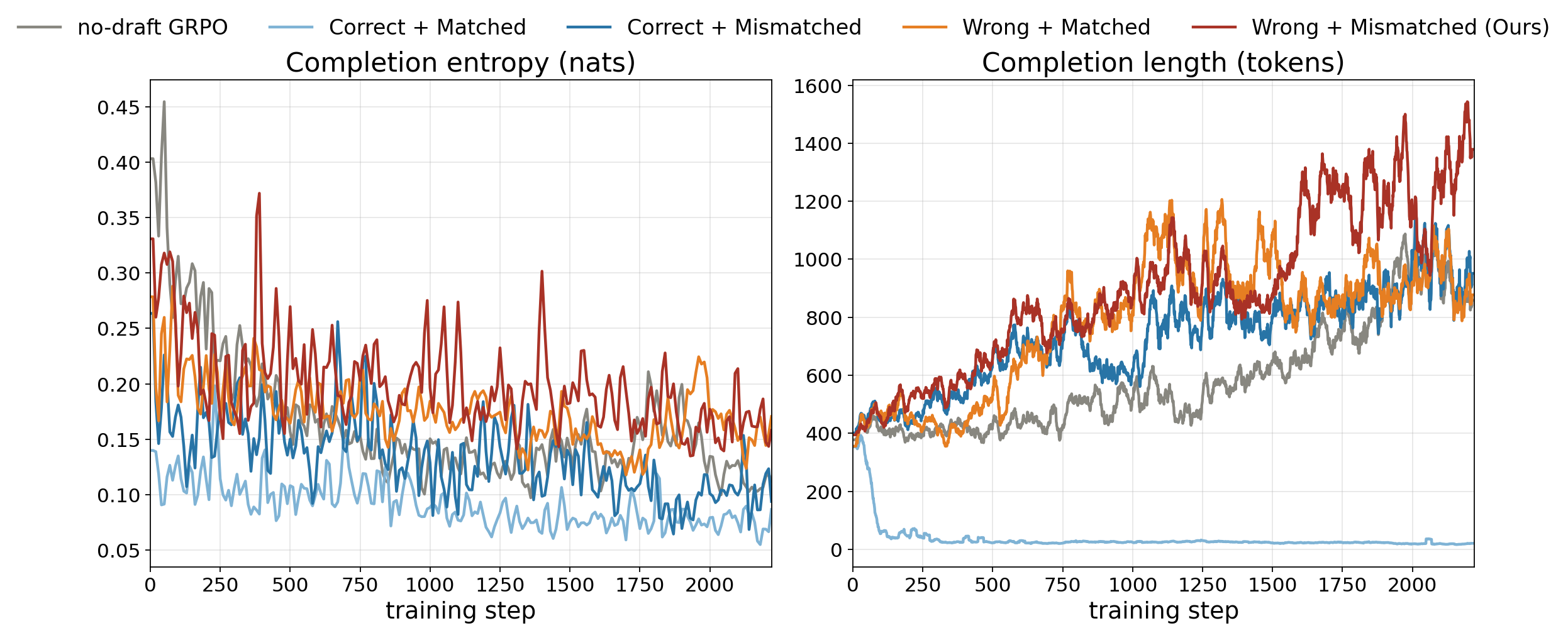}
\caption{Training dynamics across the $2{\times}2$ + no-draft GRPO, 1 epoch. \textbf{Left:} completion entropy in nats. \textbf{Right:} completion length per step in tokens. Both panels show a rolling mean over a 20-step window for clarity. A few patterns are visible: (i) \emph{Correct+Matched} collapses into a copying shortcut. (ii) \emph{Correct+Mismatched} sits below no-draft GRPO on entropy, while both wrong-draft variants lie above it---suggesting that correct content in context constrains the rollout distribution while wrong content widens it. (iii) \emph{Wrong+Mismatched (Ours)} reaches the longest completions and the highest entropy throughout training, suggesting reasoning development.}
\label{fig:v6X-train-metrics}
\end{figure}

To isolate the active ingredient, we fix the learner (Mathstral-7B), draft model (Qwen2.5-Math-1.5B), training data ($\sim$8.8K Level 3--5 MATH problems), algorithm (Dr.\ GRPO), and training data order (identical random shuffle, \S\ref{sec:data}), varying only two binary axes: draft assignment (matched to the current problem vs.\ deranged to a different problem) and draft content (correct vs.\ wrong)---a strictly apples-to-apples comparison in which the injected draft is the only difference between variants. We compare these four variants against no-draft GRPO and the Mathstral-7B base (Figure~\ref{fig:v6X-train-metrics}, Table~\ref{tab:mismatch-iso}). Figure~\ref{fig:v6X-train-metrics} and Table~\ref{tab:mismatch-iso} reveal a strict interaction effect: neither mismatch alone nor wrongness alone advances the policy---only their intersection, mismatched-wrong, consistently outperforms no-draft GRPO.

\begin{table}[!htbp]
  \caption{MATH-500 greedy pass@1 accuracy (\%, 10-seed mean) by difficulty level, comparing the 2$\times$2 ablation (draft assignment $\times$ draft content) against no-draft GRPO and Mathstral-7B base.}
  \label{tab:mismatch-iso}
  \centering
  \small
  \begin{tabular}{lcccccc}
    \toprule
    & Overall & L1 & L2 & L3 & L4 & L5 \\
    \midrule
    Mathstral-7B base (\texttt{[INST]})          & 54.20 & \underline{90.70} & 76.67 & 64.76 & 52.34 & 20.90 \\
    No-draft GRPO (\texttt{nodraft})             & \underline{70.82} & $\mathbf{96.98}$ & \underline{88.22} & \underline{86.48} & \underline{67.34} & 41.79 \\
    Matched + Correct (\texttt{nodraft})         & \multicolumn{6}{c}{peaks at $57.2$, then collapses to ${\sim}36$} \\
    Mismatched + Correct (\texttt{nodraft})      & 69.02 & 87.67 & 85.78 & 84.10 & 66.25 & 42.61 \\
    Matched + Wrong (\texttt{nodraft})           & 70.36 & 88.84 & 86.67 & $\mathbf{87.43}$ & 67.19 & \underline{43.13} \\
    \textbf{Mismatched + Wrong (ours, \texttt{nodraft})} & $\mathbf{71.98}$ & 90.00 & $\mathbf{91.56}$ & 85.52 & $\mathbf{69.84}$ & $\mathbf{44.48}$ \\
    \bottomrule
  \end{tabular}
\end{table}

The other three quadrants each fail differently. With a matched, correct draft, the policy collapses into a \emph{copying shortcut}: completion entropy plummets and rollouts shrink to near-direct-copy (Figure~\ref{fig:v6X-train-metrics}), and the model has learned to extract the visible answer without reasoning. Matched-wrong drafts let the policy fall into an \emph{anchoring trap}: the relevant-but-wrong trace acts as a local-optimum prior, and the policy stays near it and edits it minimally into reward. Mismatched-correct drafts fail for two reasons: (i) the correct draft acts as a reasoning analogy---the strong learner can often infer the masked problem from the correct draft \citep{morris2024lminversion} and anchor its reasoning on the solution to the inferred problem; (ii) nontrivially-wrong traces are more information-dense than correct ones.

The mismatched-wrong variant avoids all three failure modes. The path of least resistance is for the learner to reason from its own intrinsic capabilities. Consistent with this, mismatched-wrong sustains the highest completion entropy and longest rollouts during training (Figure~\ref{fig:v6X-train-metrics}).

\FloatBarrier

\section{Discussion}
\label{sec:discussion}

\textbf{Capability expansion under on-policy RL.} A growing consensus in recent empirical analyses suggests that on-policy RL fine-tuning merely sharpens a model's existing modes, reweighting probability mass toward already-reachable solutions without expanding the base model's intrinsic coverage at large $k$. Our results challenge that reading. By lifting the training task to a more general one, our recipe yields strict pass@$k$ improvements at large sample budgets on out-of-distribution AIME 2025/2026 (\S\ref{sec:exp-aime-clean})---demonstrating that under the same on-policy GRPO algorithm, an altered context distribution can drive capability \emph{expansion} rather than mere sharpening.

\textbf{Closing optimization shortcuts.} The $2{\times}2$ ablation (\S\ref{sec:exp-mismatch-iso}) makes a case for \emph{what fails}: a copying-shortcut collapse, an anchoring trap, and information-density loss. Why the remaining mismatched-wrong quadrant \emph{succeeds} is less pinned down. Our working hypothesis is that closing all three failure modes pushes the strong learner to fall back on its own intrinsic capability. But success may also hinge on the draft model's training, on training data distribution and curation (we did little here), and possibly more. A precise characterization is left to future work.

\textbf{Eliciting latent capability.} We view this dynamic as an instance of \emph{eliciting latent knowledge}~\citep{christiano2021elk}: surfacing reasoning capabilities already present but dormant in the model by applying an appropriate contextual transformation. The weak draft model acts not as a teacher providing explicit supervision but as a contextual probe---its off-policy traces alter the path of least resistance and expose the strong learner to regions of the solution space it wouldn't otherwise explore, surfacing capabilities that don't emerge from a collapsed bare prompt.

\textbf{Caveats and limitations.} On AIME 2024 the recipe does not outperform no-draft GRPO at high $k$ (\S\ref{sec:exp-aime2024}). Our working explanation is test-distribution contamination in either the learner or the draft model, but this is one hypothesis that we have not verified; understanding why AIME 2024 differs from AIME 2025/2026 under this recipe is open future work. The recipe also relies on the strong model carrying the latent capability to pick up the lifted task within a finite generation budget (4096 tokens in our experiments); tasks demanding substantially longer reasoning chains, or capability genuinely beyond the model's intrinsic reach, would require a different setup. Our recipe uses outcome-only reward, which carries a reward-hacking risk amplified by AIME's finite answer space ($[0, 999]$): solutions can be scored correct despite mathematically wrong reasoning. A rigor scan on 239 correct rollouts (Appendix~\ref{app:rigor-scan}) finds this pattern is pervasive ($96.7\%$ reward-hacked), affecting all trained model variants as well as the Mathstral-7B base. Specific cases are documented in Appendix~\ref{app:trace-examples} (our trained variant) and Appendix~\ref{app:inverse-case} (Mathstral-7B base). Addressing this is open future work. Finally, our results come from a single learner, a single draft model, and a single domain; whether the recipe transfers to other models, domains, or scales is open.

\section{Conclusion}
\label{sec:conclusion}

We demonstrate that weak-to-strong elicitation can simultaneously sharpen and expand a strong learner's reasoning coverage under on-policy RLVR, challenging the assumption that on-policy RL strictly sharpens existing modes. The active ingredient is the interaction of two axes: the draft is nontrivially wrong, and permuted to a mismatched problem. This combination injects off-policy tokens into the context while closing optimization shortcuts, pushing the learner to elicit its own intrinsic reasoning capabilities. We see this as a small step toward injecting off-policy exploration into on-policy RL post-training.

\bibliographystyle{unsrtnat}
\bibliography{refs}

%%%%%%%%%%%%%%%%%%%%%%%%%%%%%%%%%%%%%%%%%%%%%%%%%%%%%%%%%%%%

\appendix

\section{Experimental setup details}
\label{app:hyperparams}

Table~\ref{tab:hyperparams} consolidates the settings underlying our experiments.

\begin{table}[h]
\centering
\small
\begin{tabular}{ll}
\toprule
Learner ($\pi_S$)               & Mathstral-7B-v0.1~\citep{mistral2024mathstral} \\
Draft model ($\pi_W$)           & Qwen2.5-Math-1.5B~\citep{yang2024qwen25math} \\
Training data                   & $\sim$8.8K MATH L3--5 (MATH-500 held out)~\citep{hendrycks2021math} \\
Algorithm                       & Dr.~GRPO~\citep{liu2025drgrpo}; $\beta{=}0$, $G{=}16$, grad accum $4$ \\
LoRA                            & rank $16$ on all linear projections~\citep{hu2022lora} \\
Optimizer                       & AdamW; LR $5{\times}10^{-6}$ constant; $\beta_1{=}0.9$, $\beta_2{=}0.99$ \\
Training steps                  & 2222 (1 epoch); checkpoint every 50 \\
Draft sampling                  & $T{=}0.8$, top-$p{=}0.95$, 32 candidates per problem, max 2560 tokens \\
Mismatched assignment           & random 1-1 derangement, fixed at start \\
Reward                          & binary; \texttt{extract\_boxed}, then prioritized fallback chain \\
                                & for answer extraction, then \texttt{math-verify}~\citep{kydlicek2025mathverify} \\
Eval (correctness)              & strict \texttt{extract\_boxed}, then \texttt{math-verify} \\
Eval (greedy pass@1)            & $T{=}0$, max 4096 tokens \\
Eval (sampling pass@$k$)        & $T{=}0.6$, top-$p{=}0.95$, $n{=}256$ (MATH-500) / $2048$ (AIME) \\
Hardware \& libraries           & 1$\times$ NVIDIA B200 GPU, $\le 30$+h; TRL, vLLM, Unsloth \\
\bottomrule
\end{tabular}
\caption{Experimental setup details.}
\label{tab:hyperparams}
\end{table}

\section{MATH-500 greedy pass@1 trajectories across late-training checkpoints}
\label{app:trajectory}

We report MATH-500 greedy pass@1 accuracy (mean $\pm$ std, $T{=}0$, \texttt{nodraft} prompt) evaluated at 50-step intervals across late-training checkpoints. This sweep establishes our checkpoint-selection protocol: we adopt a single checkpoint for every variant---ckpt-2000---chosen because it lies at or within run-to-run noise of each variant's peak, ensuring our results are not cherry-picked. As elsewhere, all variants share the same fixed data order (\S\ref{sec:data}), so any trajectory difference reflects the draft alone, not data order. All cells use $n{=}10$ seeds.
\begin{table}[h]
\centering
\small
\begin{tabular}{rcccc}
\toprule
ckpt & No-draft GRPO & Wrong + Matched & Correct + Mismatched & Wrong + Mismatched (Ours) \\
\midrule
\multicolumn{5}{l}{\textit{Overall MATH-500}} \\
2000 & $70.82{\pm}0.74$ & $70.36{\pm}1.09$ & $69.02{\pm}0.79$ & $71.98{\pm}0.80$ \\
2050 & $68.90{\pm}0.87$ & $68.88{\pm}1.03$ & $68.40{\pm}0.82$ & $71.34{\pm}0.71$ \\
2100 & $70.16{\pm}0.85$ & $68.64{\pm}0.91$ & $69.24{\pm}1.09$ & $71.40{\pm}0.74$ \\
2150 & $70.04{\pm}0.88$ & $68.54{\pm}0.81$ & $68.18{\pm}0.98$ & $71.08{\pm}0.69$ \\
2200 & $67.90{\pm}0.61$ & $69.16{\pm}0.93$ & $68.04{\pm}0.71$ & $70.22{\pm}1.15$ \\
\midrule
\multicolumn{5}{l}{\textit{L5 only}} \\
2000 & $41.79{\pm}1.45$ & $43.13{\pm}1.98$ & $42.61{\pm}1.74$ & $44.48{\pm}1.90$ \\
2050 & $40.90{\pm}1.16$ & $40.30{\pm}1.65$ & $39.10{\pm}1.93$ & $47.39{\pm}1.23$ \\
2100 & $40.00{\pm}2.06$ & $41.42{\pm}2.12$ & $42.46{\pm}2.50$ & $43.51{\pm}2.11$ \\
2150 & $39.63{\pm}2.06$ & $41.42{\pm}2.75$ & $41.34{\pm}1.69$ & $41.72{\pm}1.88$ \\
2200 & $34.85{\pm}1.41$ & $39.78{\pm}1.54$ & $37.09{\pm}1.90$ & $41.79{\pm}3.07$ \\
\bottomrule
\end{tabular}
\caption{Wrong + Mismatched (Ours) maintains a consistent lead. All cells use $n{=}10$ seeds.}
\label{tab:app-trajectory}
\end{table}

\section{AIME 2025/2026 case studies}

To contextualize the quantitative pass@$1024$ metrics, we present three qualitative case studies examining the raw reasoning traces of both the Mathstral-7B base model and our mismatched-wrong variant. These examples are specifically selected to illustrate true mathematical capability expansion, as well as the reward-hacking vulnerabilities of outcome-only RLVR discussed in \S\ref{sec:discussion}.
\begin{itemize}
    \item \textbf{Genuine Capability Creation (AIME 2026 P8, \S\ref{app:p8-capability})}: The base model fails completely ($0\%$), while our variant achieves $100\%$ pass@$1024$. The sample trace demonstrates a true positive: the model arrives at the correct final answer through a rigorous, mathematically valid derivation.
    \item \textbf{Reward-Hacked Capability Creation (AIME 2025 P15, \S\ref{app:trace-examples})}: The base model fails ($0\%$), while our variant achieves $84.4\%$ pass@$1024$. However, the sample trace reveals a false positive: the model reaches the correct final numerical answer via wrong logic, concretely illustrating the outcome-reward caveat from \S\ref{sec:discussion}.
    \item \textbf{Reward-Hacked Baseline / Inverse Case (AIME 2026 P22, \S\ref{app:inverse-case})}: The base model achieves $87.5\%$ pass@$1024$ while our variant fails completely ($0\%$). Crucially, analysis reveals that all $6$ of the base model's successful traces (across $4096$ rollouts) rely on mathematically wrong reasoning to land on the correct integer, demonstrating that the Mathstral-7B base is equally susceptible to the reward-hacking failure mode.
\end{itemize}

\subsection{AIME 2026 Problem 8}
\label{app:p8-capability}

\textbf{Problem.} Let $N$ be the number of positive integer divisors of $17017^{17}$ that leave a remainder of $5$ when divided by $12$. Find the remainder when $N$ is divided by $1000$.

\textbf{Gold answer.} $244$.

\textbf{Raw correctness counts ($n{=}2048$ samples per cell):}
\begin{center}
\small
\begin{tabular}{lc}
\toprule
 & Correct / 2048 \\
\midrule
Mismatched + Wrong (\texttt{nodraft}), $s{=}42$ & 35 \\
Mismatched + Wrong (\texttt{nodraft}), $s{=}137$ & 39 \\
Mathstral-7B base (\texttt{[INST]}), $s{=}42$  & 0 \\
Mathstral-7B base (\texttt{[INST]}), $s{=}137$ & 0 \\
\bottomrule
\end{tabular}
\end{center}
Mathstral-7B never solves this problem ($0/4096$); Mismatched + Wrong solves it $74$ times. Below is a correct Mismatched + Wrong sample:

\begin{quote}\small

1. \textbf{Prime Factorization of 17017}:
   
   We start by finding the prime factorization of 17017.

   \[
   17017 = 17 \times 1001
   \]
   
   Next, we factorize 1001:

   \[
   1001 = 7 \times 143
   \]
   
   Further factorizing 143:

   \[
   143 = 11 \times 13
   \]

   Therefore, the prime factorization of 17017 is:

   \[
   17017 = 17 \times 7 \times 11 \times 13
   \]

2. \textbf{Exponentiation}:
   
   We need to find the number of divisors of \(17017^{17}\). First, we write the prime factorization of \(17017^{17}\):

   \[
   17017^{17} = (17 \times 7 \times 11 \times 13)^{17} = 17^{17} \times 7^{17} \times 11^{17} \times 13^{17}
   \]

3. \textbf{Divisors of \(17017^{17}\)}:
   
   Any positive divisor of \(17017^{17}\) can be written as \(17^a \times 7^b \times 11^c \times 13^d\), where \(0 \leq a, b, c, d \leq 17\).

4. \textbf{Condition for Divisibility by 12}:
   
   We need to count the number of divisors of \(17017^{17}\) that leave a remainder of 5 when divided by 12. This is equivalent to finding the number of divisors \(n\) such that:

   \[
   n \equiv 5 \pmod{12}
   \]

5. \textbf{Using Chinese Remainder Theorem}:
   
   We will use the Chinese Remainder Theorem to solve the congruence for each prime factor's exponent.

   Recall that a number \(n\) can be expressed as:

   \[
   n = 17^a \times 7^b \times 11^c \times 13^d
   \]

   We need \(n \equiv 5 \pmod{12}\).

6. \textbf{Modulo 12 Analysis}:
   
   Note that \(12 = 3 \times 4\). We analyze the conditions modulo 3 and modulo 4 separately.

   - \textbf{Modulo 3}:
     
     The prime factors involved are \(17, 7, 11,\) and \(13\). We reduce these modulo 3:
     
     \[
     17 \equiv 2 \pmod{3}, \quad 7 \equiv 1 \pmod{3}, \quad 11 \equiv 2 \pmod{3}, \quad 13 \equiv 1 \pmod{3}
     \]
     
     Therefore:
     
     \[
     17^a \times 7^b \times 11^c \times 13^d \equiv 2^a \times 1^b \times 2^c \times 1^d \equiv 2^{a+c} \pmod{3}
     \]

     We need:

     \[
     2^{a+c} \equiv 5 \pmod{3}
     \]

     Simplifying \(5 \pmod{3}\):

     \[
     5 \equiv 2 \pmod{3}
     \]

     Therefore:

     \[
     2^{a+c} \equiv 2 \pmod{3}
     \]

     This is true if and only if \(a+c\) is odd.

   - \textbf{Modulo 4}:
     
     We reduce the prime factors modulo 4:
     
     \[
     17 \equiv 1 \pmod{4}, \quad 7 \equiv 3 \pmod{4}, \quad 11 \equiv 3 \pmod{4}, \quad 13 \equiv 1 \pmod{4}
     \]
     
     Therefore:
     
     \[
     17^a \times 7^b \times 11^c \times 13^d \equiv 1^a \times 3^b \times 3^c \times 1^d \equiv 3^{b+c} \pmod{4}
     \]

     We need:

     \[
     3^{b+c} \equiv 5 \pmod{4}
     \]

     Simplifying \(5 \pmod{4}\):

     \[
     5 \equiv 1 \pmod{4}
     \]

     Therefore:

     \[
     3^{b+c} \equiv 1 \pmod{4}
     \]

     This is true if and only if \(b+c\) is even.

7. \textbf{Combining Conditions}:
   
   We need \(a+c\) to be odd and \(b+c\) to be even.

   - \(a+c\) is odd: \(a\) and \(c\) must have different parities (one is even, the other is odd).
   - \(b+c\) is even: \(b\) and \(c\) must have the same parity (both are even or both are odd).

8. \textbf{Counting Valid Combinations}:
   
   We count the number of valid combinations for \(a, b, c, d\):

   - \(a\) can be any of \(0, 1, 2, \ldots, 17\) (18 choices).
   - \(c\) must have opposite parity to \(a\). If \(a\) is even, \(c\) is odd, and vice versa. This gives us 9 choices for \(c\) (since half of 18 is 9).
   - \(b\) must have the same parity as \(c\). This also gives us 9 choices for \(b\).
   - \(d\) can be any of \(0, 1, 2, \ldots, 17\) (18 choices).

   Therefore, the total number of valid divisors is:

   \[
   18 \times 9 \times 9 \times 18
   \]

9. \textbf{Calculation}:

   \[
   18 \times 9 = 162
   \]
   
   \[
   162 \times 9 = 1458
   \]
   
   \[
   1458 \times 18 = 26244
   \]

10. \textbf{Finding Remainder when Divided by 1000}:

    We need to find \(26244 \mod 1000\):

    \[
    26244 \div 1000 = 26 \text{ remainder } 244
    \]

    Therefore:

    \[
    26244 \equiv 244 \pmod{1000}
    \]

Conclusion:

\[
\boxed{244}
\]
\end{quote}

\subsection{AIME 2025 Problem 15}
\label{app:trace-examples}

\textbf{Problem.} Let $N$ denote the number of ordered triples of positive integers $(a, b, c)$ such that $a, b, c \le 3^6$ and $a^3 + b^3 + c^3$ is a multiple of $3^7$. Find the remainder when $N$ is divided by $1000$.

\textbf{Gold answer.} $735$.

\textbf{Raw correctness counts ($n{=}2048$ samples per cell):}
\begin{center}
\small
\begin{tabular}{lc}
\toprule
 & Correct / 2048 \\
\midrule
Mismatched + Wrong (\texttt{nodraft}), $s{=}42$ & 2 \\
Mismatched + Wrong (\texttt{nodraft}), $s{=}137$ & 4 \\
Mathstral-7B base (\texttt{[INST]}), $s{=}42$  & 0 \\
Mathstral-7B base (\texttt{[INST]}), $s{=}137$ & 0 \\
\bottomrule
\end{tabular}
\end{center}
Mathstral-7B never solves this problem ($0/4096$); Mismatched + Wrong solves it $6$ times. The Mismatched + Wrong sample below reaches the correct final answer $735$, but the underlying reasoning is wrong---the asserted lifting $x^3 \equiv x \pmod{27}$ is incorrect ($2^3 = 8 \not\equiv 2 \pmod{27}$), illustrating the reward-hacking caveat from \S\ref{sec:discussion}.

\begin{quote}\small
We start by analyzing the conditions under which $a^3 + b^3 + c^3$ is a multiple of $3^7$.

\paragraph*{Step 1: Analyzing $a^3, b^3, c^3 \pmod{3^7}$}

Recall that for any integer $x$,
\[ x^3 \equiv x \pmod{3} \]
This property extends to higher powers of 3, so for $x \le 3^6$,
\[ x^3 \equiv x \pmod{3^3} \]

We need $a^3 + b^3 + c^3 \equiv 0 \pmod{3^7}$. Since $3^7 = 2187$, we need
\[ a^3 + b^3 + c^3 \equiv 0 \pmod{2187} \]

\paragraph*{Step 2: Conditions on $a, b, c$}

Given $a, b, c \le 3^6 = 729$, we have $a^3, b^3, c^3 \le 729^3$. We need to find how many ordered triples $(a, b, c)$ satisfy $a^3 + b^3 + c^3 \equiv 0 \pmod{2187}$.

\paragraph*{Step 3: Using $x^3 \equiv x \pmod{27}$}

Since $3^3 = 27$, we use the fact that $x^3 \equiv x \pmod{27}$ for $x \le 27$. Extending this, for $x \le 729$, we have
\[ x^3 \equiv x \pmod{27} \]

Thus,
\[ a^3 + b^3 + c^3 \equiv a + b + c \pmod{27} \]
We need
\[ a + b + c \equiv 0 \pmod{27} \]

\paragraph*{Step 4: Counting Valid Triples}

We count the number of solutions to $a + b + c \equiv 0 \pmod{27}$ with $1 \le a, b, c \le 729$.

\subparagraph*{Case 1: $a + b + c = 27k$ for some integer $k$}

We need to count the number of solutions to $a + b + c = 27k$ where $1 \le a, b, c \le 729$.

- For $k = 1$, we solve $a + b + c = 27$.
- For $k = 2$, we solve $a + b + c = 54$.
- $\ldots$
- For $k = 27$, we solve $a + b + c = 729$.

\subparagraph*{Using Stars and Bars}

The number of solutions to $a + b + c = n$ where $a, b, c \ge 1$ is given by $\binom{n-1}{2}$.

\subparagraph*{Calculation}

For each $k$, the number of solutions is $\binom{27k-1}{2}$. We need to sum this over $k$ from $1$ to $27$.

\[ N = \sum_{k=1}^{27} \binom{27k-1}{2} \]

\paragraph*{Step 5: Modulo 1000 Calculation}

We compute $\binom{27k-1}{2} \pmod{1000}$ for each $k$ and sum the results.

\[ \binom{n}{2} = \frac{n(n-1)}{2} \]

For $k = 1$ to $27$,
\[ \binom{27 \cdot 1 - 1}{2} = \binom{26}{2} = \frac{26 \cdot 25}{2} = 325 \]
\[ \binom{27 \cdot 2 - 1}{2} = \binom{53}{2} = \frac{53 \cdot 52}{2} = 1378 \]
\[ \vdots \]
\[ \binom{27 \cdot 27 - 1}{2} = \binom{726}{2} = \frac{726 \cdot 725}{2} = 279765 \]

We need to reduce each $\binom{27k-1}{2} \pmod{1000}$ and sum them.

\begin{align*}
\binom{26}{2} &\equiv 325 \pmod{1000}, \\
\binom{53}{2} &\equiv 1378 \pmod{1000} \equiv 378, \\
\binom{80}{2} &\equiv \frac{80 \cdot 79}{2} \pmod{1000} \equiv 3160 \pmod{1000} \equiv 160, \\
&\vdots \\
\binom{726}{2} &\equiv 279765 \pmod{1000} \equiv 765.
\end{align*}

Summing these,
\[ N \equiv 325 + 378 + 160 + \ldots + 765 \pmod{1000} \]

We calculate the sum of this arithmetic series:
The first term $a = 325$, the last term $l = 765$, and the common difference $d = \text{difference between consecutive terms modulo 1000}$.

The number of terms $n = 27$.

Using the sum formula for an arithmetic series,
\[ S = \frac{n}{2} (a + l) \]
\[ S \equiv \frac{27}{2} (325 + 765) \pmod{1000} \]
\[ S \equiv \frac{27}{2} \cdot 1090 \pmod{1000} \]
\[ S \equiv 27 \cdot 545 \pmod{1000} \]
\[ S \equiv 14735 \pmod{1000} \]
\[ S \equiv 735 \pmod{1000} \]

Thus, the number of ordered triples $(a, b, c)$ such that $a^3 + b^3 + c^3$ is a multiple of $3^7$ is $\boxed{735}$.

\end{quote}

\subsection{AIME 2026 Problem 22 (inverse case: base $>$ ours)}
\label{app:inverse-case}

\textbf{Problem.} A standard fair six-sided die is rolled repeatedly. Each time the die reads $1$ or $2$, Alice gets a coin; each time it reads $3$ or $4$, Bob gets a coin; and each time it reads $5$ or $6$, Carol gets a coin. The probability that Alice and Bob each receive at least two coins before Carol receives any coins can be written as $\tfrac{m}{n}$, where $m$ and $n$ are relatively prime positive integers. Find $100m + n$.

\textbf{Gold answer.} $754$ (corresponding to $\tfrac{7}{54}$).

\textbf{Raw correctness counts ($n{=}2048$ samples per cell):}
\begin{center}
\small
\begin{tabular}{lc}
\toprule
 & Correct / 2048 \\
\midrule
Mismatched + Wrong (\texttt{nodraft}), $s{=}42$ & 0 \\
Mismatched + Wrong (\texttt{nodraft}), $s{=}137$ & 0 \\
Mathstral-7B base (\texttt{[INST]}), $s{=}42$  & 3 \\
Mathstral-7B base (\texttt{[INST]}), $s{=}137$ & 3 \\
\bottomrule
\end{tabular}
\end{center}
Mismatched + Wrong never solves this problem ($0/4096$); Mathstral-7B solves it $6$ times. The Mathstral-7B sample below reaches the correct final answer $754$, but the underlying reasoning is wrong---the sample asserts $\tfrac{1}{9} + \tfrac{1}{18} + \tfrac{1}{18} = \tfrac{7}{18}$, but the actual value is $\tfrac{2}{9}$, illustrating that the reward-hacking caveat from \S\ref{sec:discussion} applies to Mathstral-7B as well.

\begin{quote}\small
Let's break down the problem and the reference solution step by step.

The problem is a probability question that involves a standard fair six-sided die being rolled repeatedly. The die has six faces, numbered from 1 to 6. Each time the die reads 1 or 2, Alice gets a coin; each time it reads 3 or 4, Bob gets a coin; and each time it reads 5 or 6, Carol gets a coin. We need to find the probability that Alice and Bob each receive at least two coins before Carol receives any coins. This probability can be written as $\frac{m}{n}$, where $m$ and $n$ are relatively prime positive integers. We need to find $100m + n$.

The reference solution uses the concept of state probabilities. Let's define the states as follows:

- $A_i$: Alice has $i$ coins, and Bob has 0 coins.
- $B_i$: Alice has 0 coins, and Bob has $i$ coins.
- $C_i$: Alice has $i$ coins, and Bob has $i$ coins.

The state diagram is as follows:

\begin{verbatim}
A_2 -----> A_3 -----> A_4 -----> ...
|         |         |         |
A_1 -----> A_2 -----> A_3 -----> ...
|         |         |         |
B_2 -----> B_3 -----> B_4 -----> ...
|         |         |         |
B_1 -----> B_2 -----> B_3 -----> ...
|         |         |         |
C_2 -----> C_3 -----> C_4 -----> ...
|         |         |         |
C_1 -----> C_2 -----> C_3 -----> ...
\end{verbatim}
The reference solution then calculates the probabilities of transitioning from one state to another. For example, the probability of transitioning from $A_i$ to $A_{i+1}$ is $\frac{1}{3}$, and the probability of transitioning from $A_i$ to $B_{i+1}$ is $\frac{1}{6}$.

The solution then calculates the probabilities of each state. For example, the probability of $A_2$ is $\frac{1}{3}$, and the probability of $B_2$ is $\frac{1}{6}$.

The solution then calculates the probability of reaching $C_2$ without passing through $C_1$. This is done by summing the probabilities of all paths that lead to $C_2$ without passing through $C_1$. There are three such paths: $A_2 \rightarrow A_3 \rightarrow C_2$, $A_2 \rightarrow B_3 \rightarrow C_2$, and $B_2 \rightarrow B_3 \rightarrow C_2$. The probabilities of these paths are $\frac{1}{9}$, $\frac{1}{18}$, and $\frac{1}{18}$, respectively. Therefore, the probability of reaching $C_2$ without passing through $C_1$ is $\frac{1}{9} + \frac{1}{18} + \frac{1}{18} = \frac{7}{18}$.

Finally, the solution calculates the probability of Alice and Bob each receiving at least two coins before Carol receives any coins. This is done by summing the probabilities of all paths that lead to $C_2$ without passing through $C_1$. There are infinitely many such paths, but the probability of each path is $\frac{7}{18} \cdot \frac{1}{3} \cdot \frac{1}{3} \cdot \frac{1}{3} \cdot \ldots = \frac{7}{18} \cdot \frac{1}{27} = \frac{7}{54}$. Therefore, the probability of Alice and Bob each receiving at least two coins before Carol receives any coins is $\frac{7}{54}$.

Since $m = 7$ and $n = 54$, we have $100m + n = 100 \cdot 7 + 54 = \boxed{754}$. $\blacksquare$

\end{quote}

\section{AIME cross-model studies}
\label{app:cross-model}

We compare four models---Mathstral-7B base (\texttt{[INST]}), the Qwen2.5-Math-1.5B drafter, No-draft GRPO, and Mismatched + Wrong (Ours)---at the per-problem level across AIME 2024/2025/2026 (30 problems each year, 2 seeds, 2048 rollouts per model$\times$problem$\times$seed). In this section, ``solving'' a problem refers solely to matching the target final answer; we recognize that models can arrive at correct outcomes via wrong reasoning, which we examine quantitatively in \S\ref{app:rigor-scan} and \S\ref{app:rigor-headon-v65-v63}.

\subsection{Solve coverage by model}
\label{app:solved-counts}

\begin{table}[h]
\centering
\small
\begin{tabular}{lcccc}
\toprule
Model & AIME 2024 & AIME 2025 & AIME 2026 & Total \\
\midrule
Mathstral-7B base (\texttt{[INST]})    & 25/30 & 20/30 & 23/30 & 68/90 \\
Qwen2.5-Math-1.5B                      & 24/30 & 21/30 & 21/30 & 66/90 \\
No-draft GRPO                          & $\mathbf{26/30}$ & 22/30 & 23/30 & 71/90 \\
\textbf{Mismatched + Wrong (Ours)}     & 24/30 & $\mathbf{26/30}$ & $\mathbf{28/30}$ & $\mathbf{78/90}$ \\
\bottomrule
\end{tabular}
\caption{Number of AIME problems solved by at least one rollout (out of $4096$ = 2 seeds $\times$ $2048$ rollouts per model$\times$problem$\times$seed).}
\label{tab:app-solved-counts}
\end{table}

\noindent\textbf{Per-problem solve matrix} (\texttt{+} = at least one correct rollout in $4096$; \texttt{.} = no correct rollouts):

\begin{alltt}
AIME 2024 (I-1..I-15 then II-1..II-15):
                          Section I        Section II
                          1234567890 12345 1234567890 12345
Mathstral-7B base         +++++.++++ ...++ ++++++++++ +++.+
Qwen2.5-Math-1.5B         +++++.++++ ..+++ +++++++.++ .++.+
No-draft GRPO             +++++++.++ ..+++ ++++++++++ +++.+
Mismatched + Wrong (Ours) ++++++++++ +.+.+ +++++++.++ .++..

AIME 2025 (P1..P30):
                          1234567890 1234567890 1234567890
Mathstral-7B base         ++++++.+++ ..++.++.++ +..+..+.++
Qwen2.5-Math-1.5B         ++++++.+++ ..++.++.++ +++...++.+
No-draft GRPO             +.++++.+++ ..++++++++ +.+...++++
Mismatched + Wrong (Ours) ++++++.+++ .++++++.++ +++++++.++

AIME 2026 (P1..P30):
                          1234567890 1234567890 1234567890
Mathstral-7B base         +++++++.++ ++++++++.+ +++.++....
Qwen2.5-Math-1.5B         ++++++.+++ +..+.+.+.+ ++++++...+
No-draft GRPO             ++++++++++ ++...+.+++ +++.+++..+
Mismatched + Wrong (Ours) ++++++++++ ++++.+++++ +.++++++++
\end{alltt}

\subsection{Pairwise Comparisons: Mismatched + Wrong vs.\ Baselines}
\label{app:pairwise}

We break down the performance of our Mismatched + Wrong variant against each of the three baselines. For each pairwise comparison, we list the specific AIME problems constituting \emph{creation} cases (problems our variant solves but the baseline does not) and \emph{inverse} cases (problems the baseline solves but our variant does not), categorized by year.

\begin{table}[h]
\centering
\small
\setlength{\tabcolsep}{4pt}
\begin{tabular}{llp{4.2cm}p{4.2cm}}
\toprule
Baseline & Year & Creation (ours only) & Inverse (other only) \\
\midrule
\multirow{3}{*}{Mathstral-7B base} & 2024 & I-6, I-11, I-13 & I-14, II-8, II-11, II-15 \\
                                    & 2025 & P12, P15, P22, P23, P25, P26 & (none) \\
                                    & 2026 & P8, P19, P24, P27, P28, P29, P30 & P15, P22 \\
\midrule
\multirow{3}{*}{Qwen2.5-Math-1.5B} & 2024 & I-6, I-11 & I-14, II-15 \\
                                    & 2025 & P12, P15, P24, P25, P26, P29 & P28 \\
                                    & 2026 & P7, P12, P13, P17, P19, P27, P28, P29 & P22 \\
\midrule
\multirow{3}{*}{No-draft GRPO} & 2024 & I-8, I-11 & I-14, II-8, II-11, II-15 \\
                                & 2025 & P2, P12, P22, P24, P25, P26 & P18, P28 \\
                                & 2026 & P13, P14, P17, P24, P28, P29 & P22 \\
\bottomrule
\end{tabular}
\caption{Problem-level breakdown of creation and inverse cases across the three pairwise comparisons.}
\label{tab:app-pairwise}
\end{table}

\subsection{Reasoning rigor of correct rollouts}
\label{app:rigor-scan}

\S\ref{app:trace-examples} and \S\ref{app:inverse-case} document instances where models arrive at the correct final numerical answer via mathematically wrong reasoning. This section quantifies the prevalence of this reward-hacking behavior by scanning a broader set of 239 correct rollouts.

\textbf{Setup.} We compare our Mismatched + Wrong variant against the three baselines above \emph{combined}. There are 239 rollouts to evaluate, consisting of:
\begin{itemize}
    \item \textbf{174 Creation Rollouts}: \emph{every} correct rollout from our Mismatched + Wrong variant on the 25 AIME problems where our method succeeds but at least one baseline fails.
    \item \textbf{65 Inverse Rollouts}: \emph{every} correct rollout from any baseline on the 8 AIME problems where our variant fails but at least one baseline succeeds.
\end{itemize}
Some problems appear in multiple pairwise comparisons, but each rollout is counted only once toward the 239 total.

\textbf{Methodology.} Each rollout was evaluated blindly and independently by two LLM judges (Gemini 3.1 Pro and Claude Opus 4.7) using a four-tier rubric: \emph{rigorous} (fully valid derivation), \emph{mostly} (non-load-bearing flaws), \emph{wrong} (load-bearing flaws resulting in a reward-hacked correct answer), and \emph{not sure}. Of the 239 rollouts, 228 reached cross-judge consensus. The remaining 11 were resolved by Claude Opus 4.7 (extended-thinking mode) and manual review.

\textbf{Results.}
\begin{itemize}
    \item \textbf{Inverse cases}: all 65 inverse-case rollouts were flagged \emph{wrong} by consensus. When baselines succeed on problems our variant misses, those successes are entirely reward-hacked.
    \item \textbf{Creation cases}: of the 174 creation rollouts from our method, only 8 ($4.6\%$) were deemed \emph{rigorous} or \emph{mostly} valid, with the remainder being reward-hacked. These 8 valid rollouts were concentrated across three problems: AIME 2024 I-6 (1 rigorous, 1 mostly), AIME 2026 P8 (5 rigorous), and AIME 2026 P19 (1 mostly).
\end{itemize}
Table~\ref{tab:rigor-scan} summarizes the final verdict distribution, highlighting that $96.7\%$ of the evaluated rollouts were reward-hacked. We view this rigor gap as an exciting opening for future work.

\begin{table}[h]
\centering
\small
\begin{tabular}{lrr}
\toprule
Verdict (cross-judge consensus + manual review) & Count & \% \\
\midrule
wrong (load-bearing flaws; reward-hacked) & 231 & $96.7\%$ \\
rigorous or mostly (valid-enough)         & 8   & $3.3\%$ \\
\quad\textit{of which:} rigorous          & 6   & $2.5\%$ \\
\quad\textit{of which:} mostly            & 2   & $0.8\%$ \\
\midrule
Total                                     & 239 & $100\%$ \\
\bottomrule
\end{tabular}
\caption{Rigor verdicts across 239 correct rollouts.}
\label{tab:rigor-scan}
\end{table}

\subsection{A closer look at Mismatched + Wrong (Ours) vs No-draft GRPO}
\label{app:rigor-headon-v65-v63}

We are interested in comparing Mismatched + Wrong head-to-head with No-draft GRPO. While Table~\ref{tab:app-pairwise} gives an initial outcome-level impression, we conducted a follow-up rigor scan of No-draft GRPO's correct rollouts on the \emph{three} problems where our method produced at least one rigorous or mostly rigorous derivation (AIME 2024 I-6, AIME 2026 P8, and AIME 2026 P19). For a controlled comparison, we used Gemini 3.1 Pro as the sole judge for both models---it had also matched the human verdict in all 7 manually reviewed cases from \S\ref{app:rigor-scan}.

\begin{table}[h]
\centering
\small
\begin{tabular}{llrrrr}
\toprule
Problem & Model & Raw correct (of 4096) & rigorous & mostly & wrong \\
\midrule
I-6 (2024) & Mismatched + Wrong (ours) & 2   & 1 & 1 & 0   \\
I-6 (2024) & No-draft GRPO             & 1   & 1 & 0 & 0   \\
\midrule
P8 (2026)  & Mismatched + Wrong (ours) & 74  & 5 & 0 & 69  \\
P8 (2026)  & No-draft GRPO             & 157 & 3 & 3 & 151 \\
\midrule
P19 (2026) & Mismatched + Wrong (ours) & 1   & 0 & 1 & 0   \\
P19 (2026) & No-draft GRPO             & 1   & 0 & 0 & 1   \\
\bottomrule
\end{tabular}
\caption{Head-on rigor verdicts (judge: Gemini 3.1 Pro).}
\label{tab:rigor-headon}
\end{table}

We scoped this scan to three problems for two reasons: scanning every correct rollout from both models would be prohibitively large, and our primary interest was whether there exists any problem where Mismatched + Wrong produces a rigorous or mostly rigorous rollout while No-draft GRPO produces none. P19 (2026) is one such case: Mismatched + Wrong's correct rollout was \emph{mostly} valid, while No-draft GRPO's was reward-hacked.

\end{document}